\documentclass[sigconf,nonacm]{acmart}

\AtBeginDocument{%
  }

\setcopyright{acmlicensed}
\copyrightyear{2018}
\acmYear{2018}
\acmDOI{XXXXXXX.XXXXXXX}

\acmConference[Conference acronym 'XX]{Make sure to enter the correct
  conference title from your rights confirmation emai}{June 03--05,
  2018}{Woodstock, NY}
\acmISBN{978-1-4503-XXXX-X/18/06}

\usepackage{array, booktabs}
\usepackage{subcaption}
\newcolumntype{M}[1]{>{\centering\arraybackslash}m{#1}} 

\begin{document}

\title{AI-Driven Stylization of 3D Environments}


\author{Yuanbo Chen}
\authornote{Contributed equally to this research.}
\affiliation{%
  \institution{University of California, Berkeley}
  \city{Berkeley}
  \country{US}
}
\author{Yixiao Kang}
\authornotemark[1]
\affiliation{%
  \institution{University of California, Berkeley}
  \city{Berkeley}
  \country{US}
}
\author{Yukun Song}
\authornotemark[1]
\affiliation{%
  \institution{University of California, Berkeley}
  \city{Berkeley}
  \country{US}
}
\author{Cyrus Vachha}
\affiliation{%
  \institution{University of California, Berkeley}
  \city{Berkeley}
  \country{US}
}
\author{Sining Huang}
\affiliation{%
  \institution{University of California, Berkeley}
  \city{Berkeley}
  \country{US}
}


\begin{abstract}
  In this our system, we discuss methods to stylize a scene of 3D primitive objects into a higher fidelity 3D scene using novel 3D representations like NeRFs and 3D Gaussian Splatting. Our approach leverages existing image stylization systems and image-to-3D generative models to create a pipeline that iterativly stylizes and composites 3D objects into scenes. We show our results on adding generated objects into a scene and discuss limitations. 
\end{abstract}

\begin{teaserfigure}
  \includegraphics[width=\textwidth]{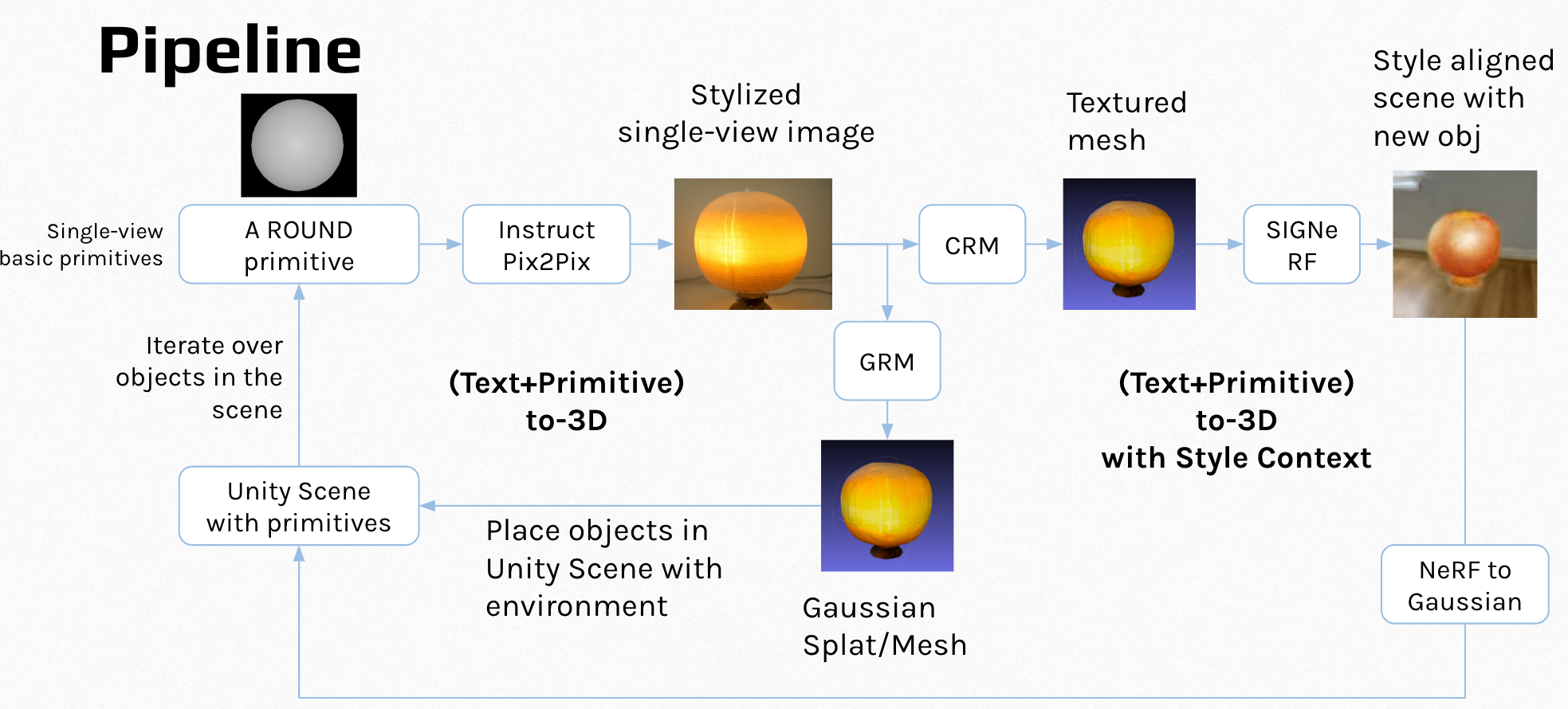}
  \caption{Overview of our method.}
  \label{fig:teaser}
\end{teaserfigure}

\maketitle

\section{Introduction}
The rapid evolution of 3D scene generation technologies has substantially enhanced the feasibility and quality of generating and manipulating three-dimensional environments, particularly through the advent of Neural Radiance Fields (NeRFs) and other generative models. These advancements have opened new possibilities for virtual reality (VR), augmented reality (AR), and other applications requiring rich, immersive, and dynamically editable 3D content. However, despite significant progress, the challenge remains in the accessibility and ease of use of these technologies for users without extensive technical expertise in 3D design. Recognizing this gap, we introduce a pipeline designed to empower individuals with little to no background in 3D design to furnish and restyle their living spaces effectively. Our approach leverages a user-friendly interface where users can draw basic primitives within a scanned room and input stylistic preferences through simple text prompts. The system then automates the conversion of these inputs into a fully furnished and stylized 3D room model, viewable in real-time.

The major contribution of our work lies in integrating systems such as InstructPix2Pix for image stylization and SIGNeRF for seamless object integration within NeRFs, tailored to simplify the design process. This pipeline not only democratizes 3D interior design but also enhances the accessibility and usability of sophisticated 3D modeling tools. By leveraging intuitive user interaction and automation, our method opens up new avenues for personal creativity and practical application in home design, offering significant benefits to users unfamiliar with traditional 3D modeling tools.

This paper outlines our methodology, discusses the implementation of our pipeline, and presents results that validate the effectiveness of our approach in creating immersive and aesthetically pleasing 3D environments tailored to user specifications. Our contribution marks a step forward in making advanced 3D scene generation accessible to a broader audience.

\section{Background}
\subsubsection{3D object generation}

Recent advances in neural representations and generative models have significantly propelled the development of 3D content generation, facilitating the creation of high-quality and diverse 3D models. The methodologies for representing 3D objects are varied and include Neural Scene Representations, Explicit Representations, Point Clouds, Meshes, Multi-layer Representations, and Implicit Representations. Among these, Neural Radiance Fields (NeRFs)~\cite{gao2022nerf} and Gaussian Splatting~\cite{kerbl20233d} are particularly notable. NeRFs, for instance, utilize a compact neural network trained to reconstruct scenes by predicting the color and intensity of light from any direction, and have rapidly gained traction in the field.

In terms of state-of-the-art technologies, the Convolutional Reconstruction Model (CRM)~\cite{wang2024crm} is noteworthy for its ability to generate six orthographic view images from a single input image, representing a significant leap in the generation of 3D models. Additionally, models like Triplane~\cite{shue20233d} and Gaussian Reconstruction Model (GRM)~\cite{xu2024grm} have shown robust performance in producing Gaussian Splatting, which is crucial for detailed scene reconstruction. These developments underscore the dynamic nature of the field and its ongoing evolution towards more sophisticated and realistic 3D content generation.

\subsection{3D Scene Generation and Editing}
Neural Radiance Fields (NeRFs) have revolutionized 3D scene reconstruction and novel view synthesis, but pose significant challenges in editing. Early initiatives like NeRF-Editing~\cite{yuan2022nerf} were limited to simpler deformations. NeuMesh~\cite{yang2022neumesh} advanced this by enabling more complex edits such as texture modifications and geometry deformation. Despite progress, current NeRF editing tools still lack the functionality of traditional 3D software and require significant manual input to achieve high-quality results.

The advent of generative NeRF editing, combining text-to-3D generation with NeRF modifications, brought new methodologies like Set-the-Scene~\cite{cohen2023set} and Compositional 3D~\cite{po2023compositional}. These methods allow for more controlled scene generation using proxy objects. More recently, Instruct-NeRF2NeRF~\cite{haque2023instruct} introduced an Iterative Dataset Update (IDU) strategy, leveraging InstructPix2Pix~\cite{brooks2023instructpix2pix} to edit NeRF datasets for more refined control over the edits.

Building on these innovations, our method uses SIGNeRF~\cite{dihlmann2024signerf}, which introduces a system that integrates edits and object generation directly within an existing NeRF scene. Using a reference sheet image grid, SIGNeRF maintains multi-view coherence and enhances control over the generation process.

\subsection{User Interaction in VR for Scene Design}
Immersive 3D modeling in Virtual Reality (VR) has transcended the spatial constraints of traditional design methods~\cite{WOLFARTSBERGER201927}. VR platforms like Dreams~\cite{dreams}, Figmin XR\cite{figmin}, and Horizon Worlds~\cite{metahorizon} have expanded user-interaction models, demonstrating VR's capability to facilitate complex design tasks through intuitive interfaces. Han et al.~\cite{Han2023-uf} further explore HCI advances in VR, focusing on enhanced gesture recognition to enable more natural and effective user interactions within virtual spaces.

\section{Method}

In this section, we present our pipeline (\autoref{fig:teaser}) for stylizing a set of basic primitives into furniture based on a user's text prompt and integrating the stylized furniture into a given scene. Our system consists of three main components: 1) a primitives stylizer, which takes a single-view image of the primitives and generates a stylized single-view image; 2) a mesh generator, which takes the stylized single-view image and generates a corresponding textured mesh; and 3) a scene integrator, which incorporates the generated mesh into the target scene. We will provide a detailed description of each component in the following subsections.

\subsection{Primitives Stylizer}
\begin{figure}[ht]
  \centering
  \includegraphics[width=\linewidth]{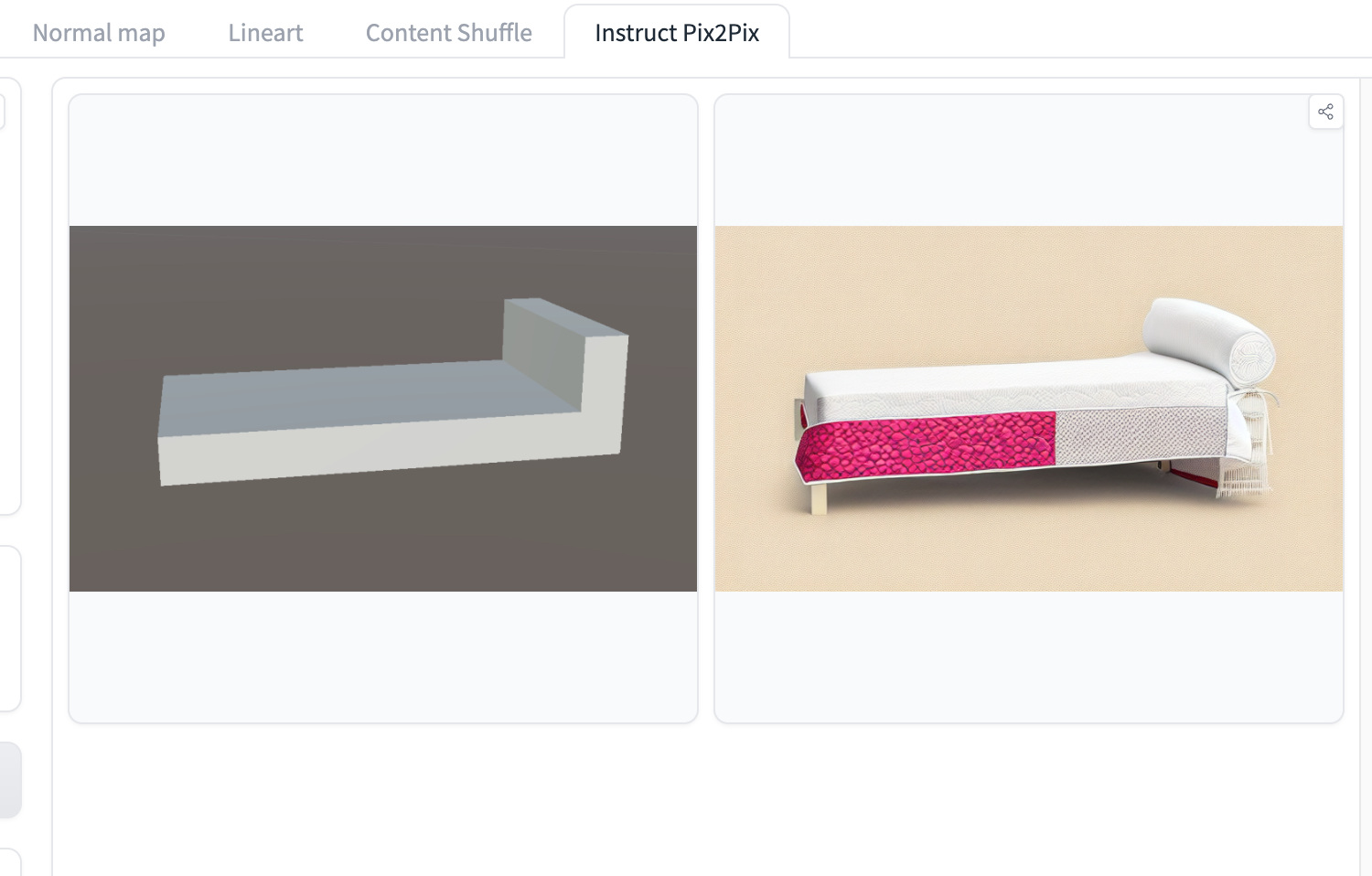}
  \caption{Result from InstructPix2Pix, with text prompt: a modern bed in the apartment, clean background}
  \label{instructPix2Pix}
\end{figure}

To stylize the set of primitives according to the user's prompt while maintaining their overall structure, we employ InstructPix2Pix~\cite{brooks2023instructpix2pix}, a state-of-the-art text-guided image editing model. InstructPix2Pix is particularly suitable for this task due to its ability to follow natural language instructions to modify specific parts of an image while preserving its overall structure and unedited regions. By leveraging the power of large-scale pre-training on a diverse set of image editing instructions, InstructPix2Pix enables us to generate high-quality stylized images that align with the user's intent while retaining the structural integrity of the primitives, as shown in Figure.\ref{instructPix2Pix}. The model's flexibility in handling a wide range of editing tasks and its capacity to generate realistic and consistent results make it an ideal choice for our primitives stylizer component.

\subsection{Mesh Generator}
After obtaining a stylized single-view image from the primitives stylizer, we aim to generate a textured mesh that maintains view consistency, minimizes photometric loss under the supervision of the given view image, and closely matches the shape of real-world objects. The underlying representation can either be surface representations (SDF, mesh) or volumetric representations (3D Gaussian blob, NeRF). In this work, we consider two state-of-the-art models for mesh generation: the Convolutional Reconstruction Model (CRM)\cite{wang2024crm} and the Gaussian Reconstruction Model (GRM)\cite{xu2024grm}.

\subsubsection{CRM}
CRM is a high-fidelity feed-forward single image-to-3D generative model that integrates geometric priors into its network design, leveraging the spatial correspondence among six orthographic images of a triplane. It first generates these orthographic view images from a single input image and then feeds them into a convolutional U-Net to create a high-resolution triplane. CRM employs Flexicubes as its geometric representation, enabling direct end-to-end optimization on textured meshes and generating high-fidelity results in about 30 seconds without test-time optimization. We choose CRM for its fast and efficient feed-forward architecture, ability to generate high-quality meshes with limited 3D data, and direct optimization on visually appealing textured meshes that closely match the stylized single-view image.

\subsubsection{GRM}
Alternatively, we also try GRM, a large-scale reconstructor capable of recovering a 3D asset from sparse-view images in around 30s. GRM is a feed-forward transformer-based model that efficiently incorporates multi-view information to translate the input pixels into pixel-aligned Gaussians, which are unprojected to create a set of densely distributed 3D Gaussians representing a scene. Together, the transformer architecture and the use of 3D Gaussians unlock a scalable and efficient reconstruction framework. Extensive experimental results demonstrate the superiority of GRM over alternatives regarding both reconstruction quality and efficiency. We also showcase the potential of GRM in generative tasks, i.e., text-to-3D and image-to-3D, by integrating it with existing multi-view diffusion models. GRM's ability to handle sparse-view inputs and its exceptional speed make it an attractive option for our mesh generator component.

\subsection{Scene Integrator}
\label{sec:signerf}
We use SIGNeRF \cite{dihlmann2024signerf} to integrate the newly generated mesh into the scene. This approach provides a technique of utilizing the ControlNet \cite{zhang2023adding} to consistently augment the existing scene across different view angles. As NeRF is trained from 2D images, the network essentially provides a mapping from stacked 2D images into the 3D environment, hence enabling the powerful 2D image editing techniques to also take effect in the 3D environment. In the SIGNeRF \cite{dihlmann2024signerf} pipeline, a set of reference camera positions are first chosen to consistently generate reference image grids, encoding color, control mask, and depth information. Then, utilizing the one slot intentionally left blank in the image grid, the SIGNeRF provides a way to iteratively update the original images in the NeRF dataset. Notice that the updating process takes the current style of NeRF scene into consideration as the pipeline utilizes ControlNet.

This approach, therefore, provides two ways to insert new objects into a NeRF scene with their styles aligned: 1) directly adding new images to the NeRF dataset and 2) modifying existing images from the NeRF dataset. However, limitations of such methods were discovered during our experiments. For the first method, the resulting NeRF dataset exhibits a spatial inconsistency across different view angles, primarily because of the lack of modification on the original dataset. On the other hand, only modifying existing images in the NeRF scene may have results highly dependent on the position chosen to insert the new object, as the original NeRF scene might not have enough camera view angles to capture sufficient information about the new objects. 
Another concern of utilizing the NeRF occurs when we iteratively add new objects to the scene. Since image editing happens on all camera view images, including both original NeRF data and the view captures generated for previously inserted objects (via approach 1), the generated view captures surrounding objects inserted earlier will accumulate higher noise compared to the augmented view captures of the latest object being inserted.
\begin{figure}[ht]
  \centering
  \includegraphics[width=\linewidth]{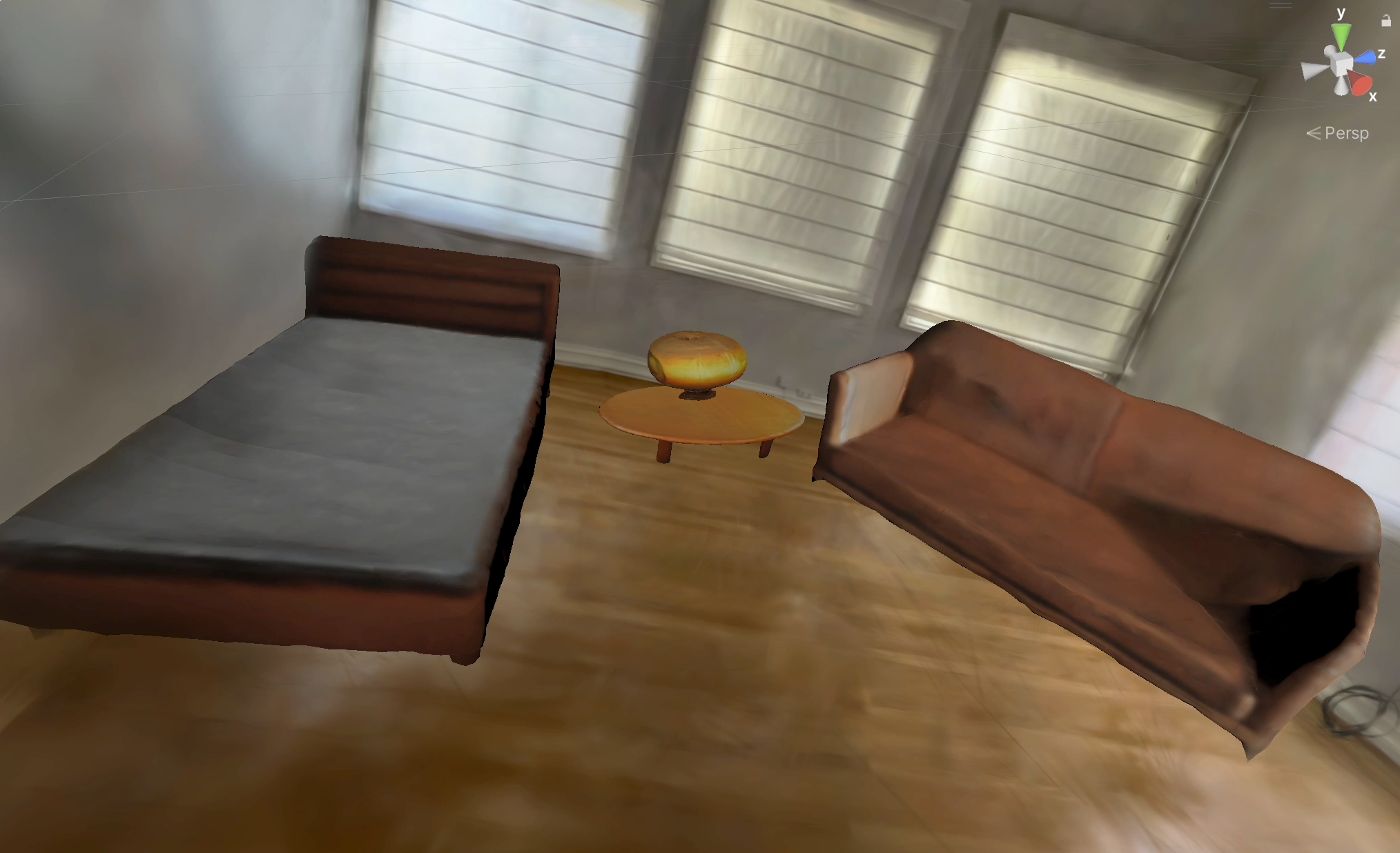}
  \caption{We combine the generated meshes from GRM in the Unity Scene}
  \label{unity}
\end{figure}
\section{Experiments}

\subsection{Experimental Setup}

We evaluated our pipeline on an empty apartment scene with three objects of different shapes and sizes that are commonly found in apartments: (1) a sofa, (2) a lamp, and (3) a bed. These three objects are iteratively added into the apartment scene at different locations, simulating how users would like to iteratively add their objects to their target scene. After adding each object, SIGNeRF retrains the scene to stylize the object according to the scene background. 

We capture the empty apartment scene with an iPhone and train an initial NeRF scene with Nerfstudio~\cite{nerfstudio}. The collected NeRF dataset has a total of 303 images and is trained in 15 mins. This scene will be used as the base for our pipeline to add objects into. Then, we utilized our proposed pipeline to add the three objects iteratively into the scene in the order of: (1) a sofa, (2) a lamp, and (3) a bed. These objects are added using the first way to add objects in the scene mentioned in section~\ref{sec:signerf}, which is to directly add new images of the object into the original NeRF dataset.

\subsection{Results}
\begin{table*}[h]
    \centering
    \begin{tabular}{@{}lcccc@{}} \toprule
        {Object} & {Primitive-Stylization (s)} & {Mesh Generation (s)} & {SIGNeRF (min)} & {Total (min)}\\ \midrule
        Sofa  & 16.7 & 30 & 28.3  & 29.1\\
        Lamp  & 18.1 & 30 & 29.1  & 29.9\\
        Bed   & 15.3 & 30 & 30.2  & 31.0\\ \bottomrule
    \end{tabular}
    \caption{Time taken for each step of the pipeline.}
    \label{tab:pipeline_time}
\end{table*}

\begin{figure*}[htb]
    \centering
    \begin{subfigure}{0.5\textwidth}
        \centering
        \includegraphics[width=0.9\linewidth]{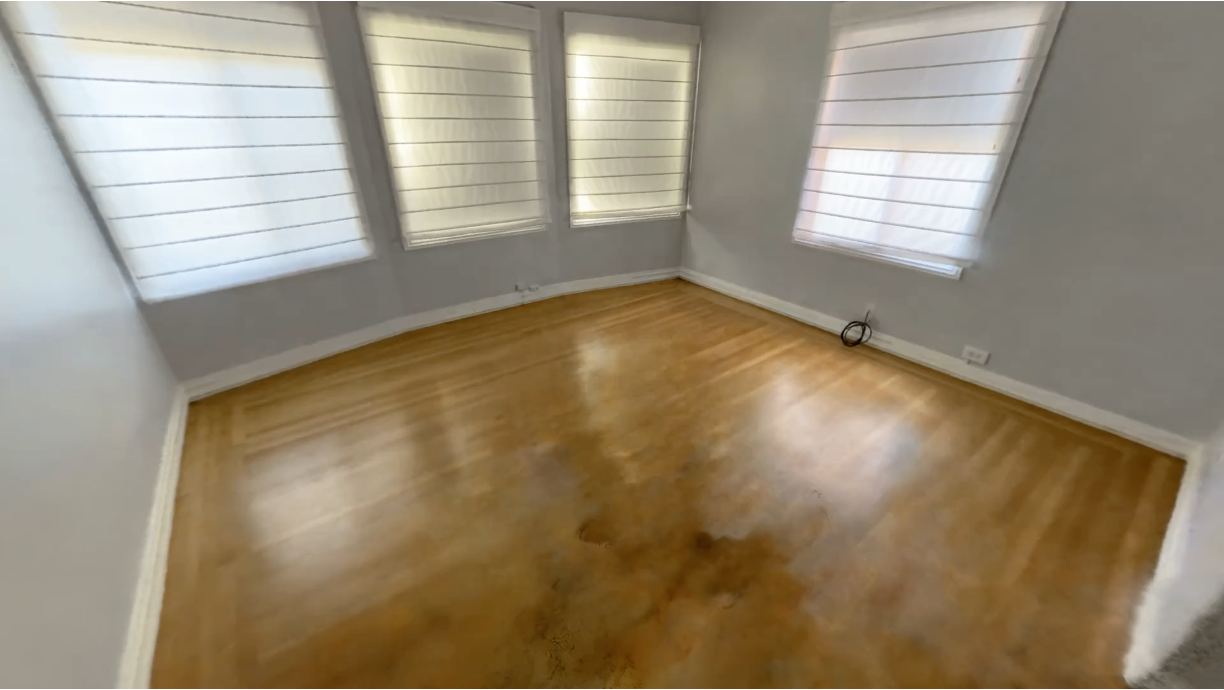}
        \caption{Initial apartment scene.}
    \end{subfigure}%
    \begin{subfigure}{0.5\textwidth}
        \centering
        \includegraphics[width=0.9\linewidth]{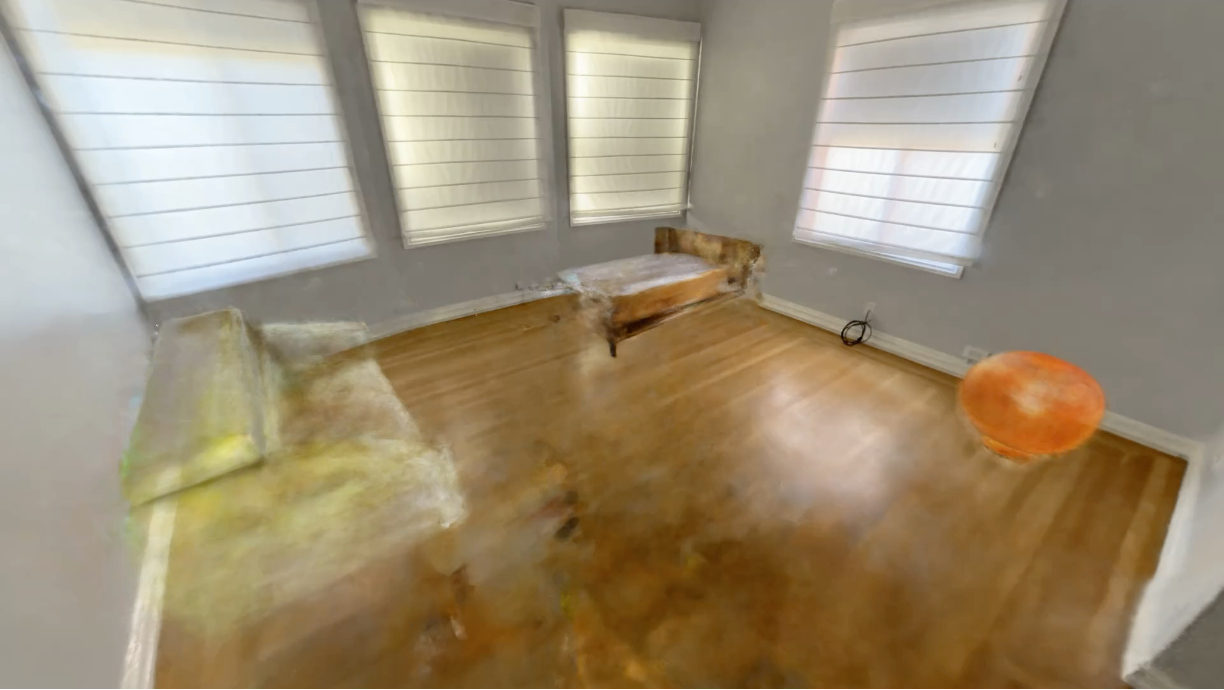}
        \caption{Apartment with sofa, bed and lamp added.}
    \end{subfigure}
    \caption{Comparison of apartment scene before and after adding objects using our pipeline. }
    \label{fig:apartment}
\end{figure*}

\begin{figure*}[htb]
\setlength\tabcolsep{3pt} 
\centering
\begin{tabular}{@{} r M{0.28\linewidth} M{0.28\linewidth} M{0.28\linewidth} @{}}
& Initial Primitive & After Stylization & Generated Mesh \\ \addlinespace
\begin{subfigure}{0.05\linewidth} \caption*{Sofa}\label{subfig:a} \end{subfigure} 
  & \includegraphics[width=\hsize]{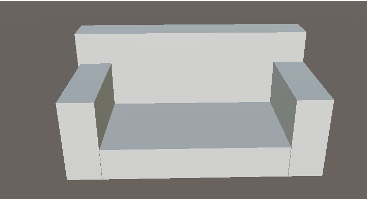} 
  & \includegraphics[width=\hsize]{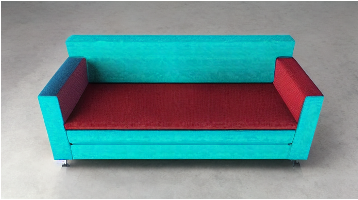}
  & \includegraphics[width=\hsize]{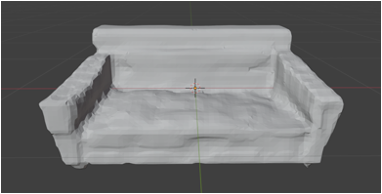}\\ \addlinespace
\begin{subfigure}{0.05\linewidth} \caption*{Lamp}\label{subfig:b} \end{subfigure} 
  & \includegraphics[width=\hsize]{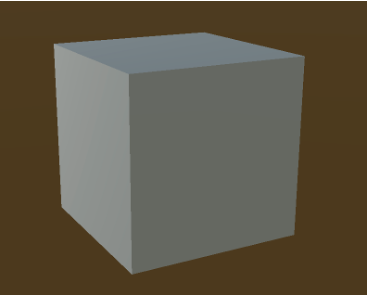}    
  & \includegraphics[width=\hsize]{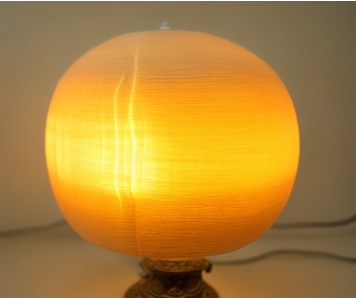}
  & \includegraphics[width=\hsize]{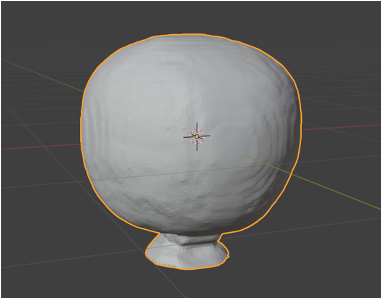}\\ \addlinespace
\begin{subfigure}{0.05\linewidth} \caption*{Bed}\label{subfig:c} \end{subfigure} 
  & \includegraphics[width=\hsize]{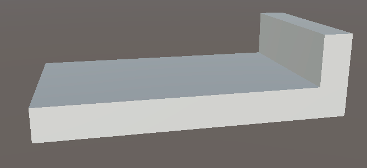} 
  & \includegraphics[width=\hsize]{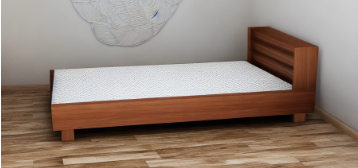}
  & \includegraphics[width=\hsize]{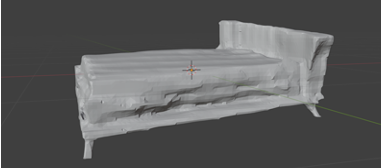}\\ \addlinespace
\end{tabular}
\caption{Results from each step in our pipeline before SIGNeRF.}
\label{fig:pre_signerf}
\end{figure*}

\begin{figure*}[htb]
\setlength\tabcolsep{3pt} 
\centering
\begin{tabular}{@{} r M{0.2\linewidth} M{0.2\linewidth} M{0.2\linewidth} M{0.2\linewidth} @{}}
\begin{subfigure}{0.05\linewidth} \caption*{Sofa}\label{subfig:a} \end{subfigure} 
  & \includegraphics[width=\hsize]{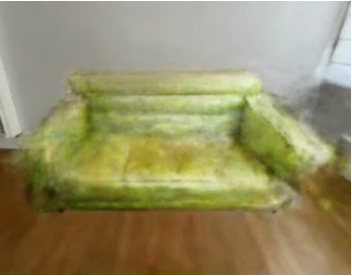} 
  & \includegraphics[width=\hsize]{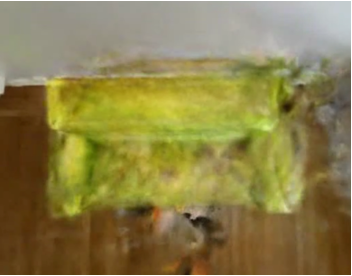}
  & \includegraphics[width=\hsize]{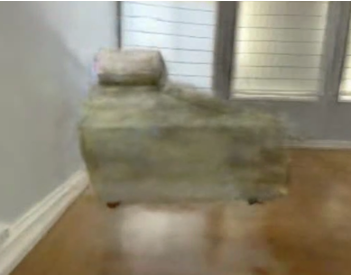}
  & \includegraphics[width=\hsize]{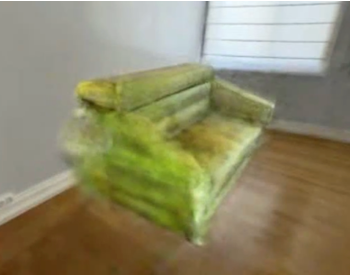}\\ \addlinespace
\begin{subfigure}{0.05\linewidth} \caption*{Lamp}\label{subfig:b} \end{subfigure} 
  & \includegraphics[width=\hsize]{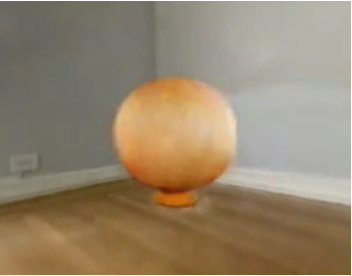}    
  & \includegraphics[width=\hsize]{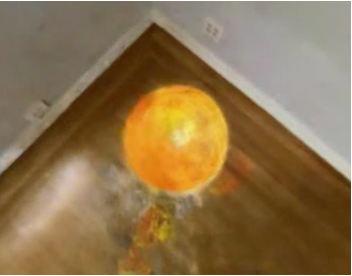}
  & \includegraphics[width=\hsize]{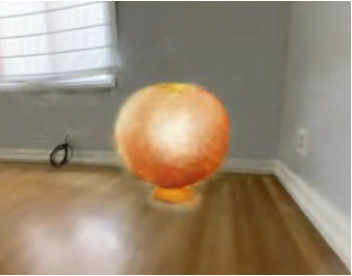}
  & \includegraphics[width=\hsize]{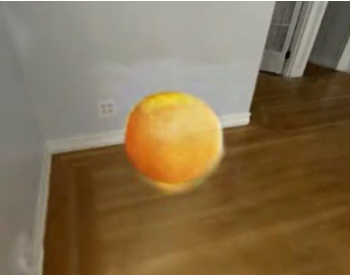}\\ \addlinespace
\begin{subfigure}{0.05\linewidth} \caption*{Bed}\label{subfig:c} \end{subfigure} 
  & \includegraphics[width=\hsize]{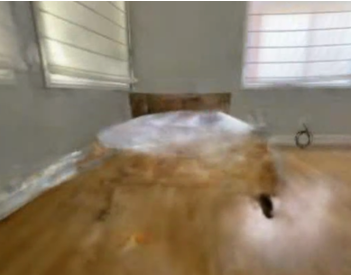} 
  & \includegraphics[width=\hsize]{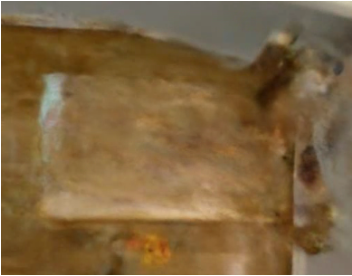}
  & \includegraphics[width=\hsize]{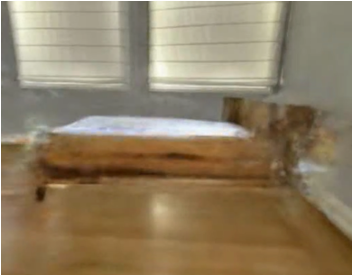}
  & \includegraphics[width=\hsize]{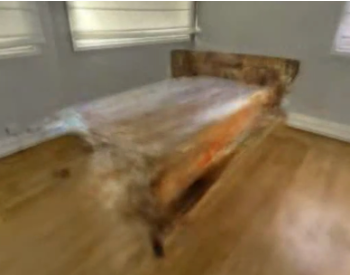}\\ \addlinespace
\end{tabular}
\vspace{-1em}
\caption{Three objects generated from our pipeline viewed at different view angles.}
\label{fig:scene_objects}
\end{figure*}

We show the before-after comparison of the apartment scene in Figure~\ref{fig:apartment}. In addition, we show the generated results for each of the three objects for each step of the entire pipeline in Figures~\ref{fig:pre_signerf} and~\ref{fig:scene_objects}. As seen in Figure~\ref{fig:apartment}, the sofa and lamp object appears translucent. This happens because they are the first two objects added into the scene. Each time we iteratively add an object into the scene, the conditioning signals such as depth and mask information of prior objects are lost. Moreover, since we utilized the first way to add objects into a scene with SIGNeRF, only the newly added images to the NeRF dataset contains the new object. The original images from the NeRF dataset are not updated, even at locations where the new object is supposed to be located. Thus, these factors contribute to the blurry and translucent results generated.

\begin{figure*}[htb]
\setlength\tabcolsep{3pt} 
\centering
\begin{tabular}{@{} r M{0.2\linewidth} M{0.2\linewidth} M{0.2\linewidth} M{0.2\linewidth} @{}}
\begin{subfigure}{0.05\linewidth} \caption*{Bed}\label{subfig:a} \end{subfigure} 
  & \includegraphics[width=\hsize]{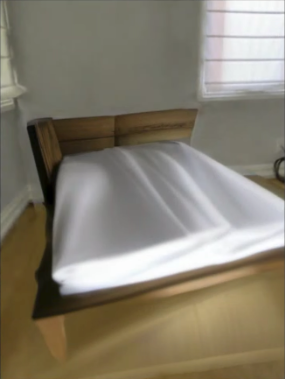} 
  & \includegraphics[width=\hsize]{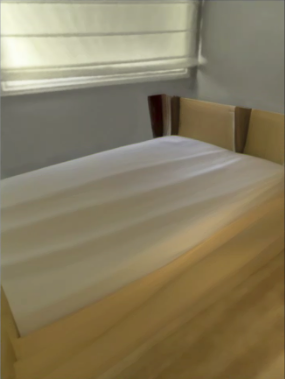}
  & \includegraphics[width=\hsize]{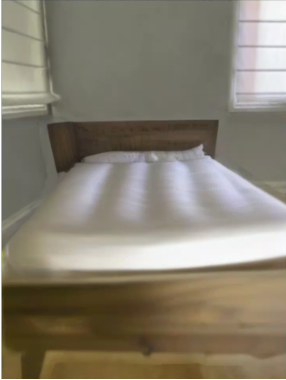}
  & \includegraphics[width=\hsize]{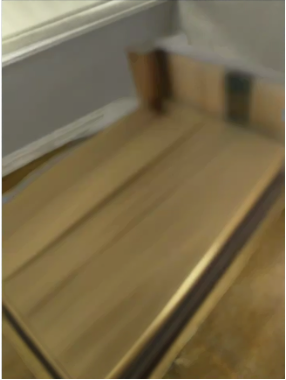}\\ \addlinespace
\begin{subfigure}{0.05\linewidth} \caption*{Sofa}\label{subfig:b} \end{subfigure} 
  & \includegraphics[width=\hsize]{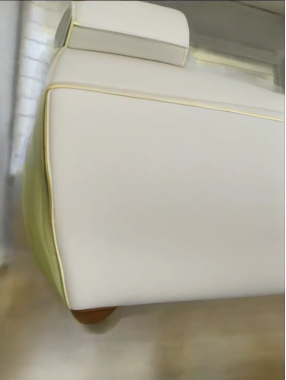}    
  & \includegraphics[width=\hsize]{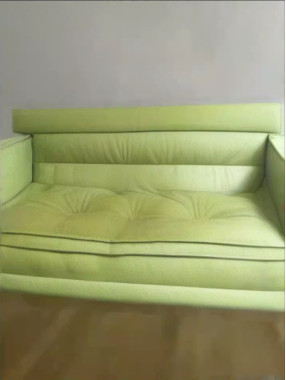}
  & \includegraphics[width=\hsize]{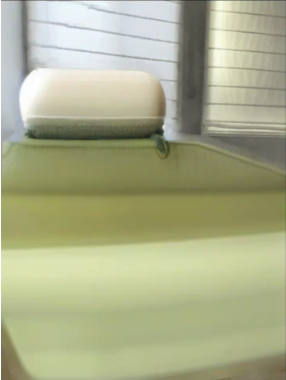}
  & \includegraphics[width=\hsize]{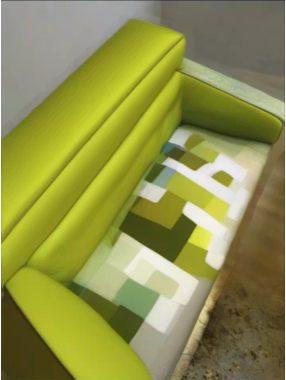}\\ \addlinespace
\end{tabular}
\vspace{-1em}
\caption{Inconsistencies of the generated dataset from SIGNeRF across different view angles.}
\label{fig:inconsistency}
\end{figure*}

From Figure~\ref{fig:scene_objects} we can observe that the objects are not consistent across different view angles, especially for the bed and for certain angles of the sofa. Upon further investigation, we discover that this is due to the inconsistencies of the generated dataset by SIGNeRF across different view angles. This can be seen from Figure~\ref{fig:inconsistency}, where the bedsheets look white from certain angles but brown with a wood-like texture from other angles. As for the sofa, it is green from most view angles but the side, which looks gray. This shows that even by conditioning the Control-Net stylization through reference grids, it is not robust enough to produce view-consistent results. 

Given the unsatisfactory blurriness and view-inconsistencies observed in the outputs from the previous pipeline, an alternative approach was implemented. The model generated by the GRM was integrated into the Unity scene, replacing the primitive forms, as depicted in Figure \ref{unity}. The resulting meshes are robust, allowing for user interactions such as moving and scaling. However, the material and color attributes of the objects, determined by the ControlNet results, do not correspond with their environmental context. This contrasts with our earlier pipeline, which stylized objects in accordance with the surrounding environment.


\section{Discussion and Limitations}
The system has a few limitations primarily for the 3D object integration and composition into the NeRF scene. In SIGNeRF, the dataset is edited with renders of composited 3D object processed through ControlNet, however, these renders are not view consistent, displaying differences in color and structure, therefore causing the resulting object to appear a bit blurry and faded at some viewpoints. Regarding the 3D object generation, the image to 3D models (CRM and GRM) sometimes show view inconsistency when generating the multi-view images and therefore the generations can appear malformed or have artifacts. This affects the quality of the stylized object. Additionally, the image stylization methods we use, Instruct-Pix2Pix and ControlNet, don't always provide significant structural changes to the input image of the primitive object beyond detailed textures. 
For future work, we propose creating an end-to-end user interface and system to allow users to easily stylize multiple 3D objects from their primitive object arrangements in Unity at once instead of iterativly (which takes much longer). We would also like to explore further methods for stylizing the background envrionment beyond taking a 2D render, stylizing with ControlNet, and generating a rough 3D scene using an image-to-3D scene system. To improve results from the image-to-3D object systems, we could find a way to leverage additional multi-view information about the primitive object instead of a single image input. To improve object composition in the NeRF scene, we could find ways to improve the view conssitency in ControlNet.

\section{Conclusion}
Overall, our system is able to demonstrate techniques for generating higher fidelity 3D objects from basic primitives for compositing into NeRF or 3DGS scenes using 2D image stylization models and image-to-3D systems. Despite current challenges in view consistency and generation quality in composited scenes, our system provides a pipeline and framework to enable future work.

\newpage
\newpage
\bibliographystyle{ACM-Reference-Format}
\bibliography{main}

\end{document}